\documentclass{article}
\usepackage{amsfonts}
\usepackage[cmex10]{amsmath}

\usepackage{times}
\usepackage{cite}
\usepackage{amsmath}
\onecolumn
\usepackage{graphicx}
\usepackage{latexsym}
\usepackage{subfigure}
\usepackage{pstricks}
\usepackage{setspace}
\date{}
\begin{document}
\title{ICABiDAS: Intuition Centred  Architecture for  Big Data Analysis and Synthesis}
\author{Amit Kumar Mishra \\
University of Cape Town\\
akmishra@ieee.org}

\footnotetext[1]{This paper is presented in the Biologically Inspired Cognitive Architecture Conference 2017 and published by their proceedings.}

\maketitle
\begin{abstract}
Humans are expert in the amount of sensory data they deal with each moment. 
Human brain not only ``analyses'' these data but also starts ``synthesizing'' new information from the existing data. 
The current age Big-data systems are needed not just to analyze data but also to come up new interpretation. 
We believe that the pivotal ability in human brain which enables us to do this is what is known as ``intuition''. 
Here, we present an intuition based architecture for big data analysis and synthesis. 
\\
\emph{\textbf{cognitive architecture, bio-inspired, intuition, synthesis, Big-Data}}
\end{abstract}

\section{Introduction}
Brain has intrigued researchers since the beginning of scientific endeavors. 
Firstly, beginning of computers saw the advent of exciting developments which culminated to the development of the new discipline of artificial neural networks (ANN). 
ANNs have been through several generations of major developments, with the recent phase consisting of spiking neural networks based works \cite{brans_99_book}. 
Another parallel field of computational neuroscience has been the bio-inspired cognitive architectures (BICA) \cite{sam_10_bica} a field which got major thrust in development. 
Cognitive architecture (CA) in general and BICA in particular also has a long history and the efforts have been devoted towards trying to emulate the functioning of brain. 
CAs like SOAR and ACT-R have been under development for many decades and have been applied in various studies \cite{rose_91_soar, taat_06_actr}. 
It may also be mentioned here that this 2012-2013 has seen multi-billion dollar investment done separately in the European Union as well as in the USA for the study and understanding of brain \cite{bluebrain, human_brain, kno_13_brain}. 

One of the use of cognitive architecture is expected to be in the emerging field of big data analytics \cite{hur_15_cog_big}. 
There are some proposed models which try to achieve this \cite{thomson2014human}. However, most of these works are still in their infancy.  
Another problem we see in the existing approaches is also the zeal to use generalized cognitive architectures. 
Generalized cognitive architectures are complex and unless we fine-tune them properly we may not achieve the desired goals. 
Lastly, there is a lack of properly defined test-cases to test any developed system. 

In our current work, we take intuition as that human trait which, we believe, can be studied to understand much of human ability at comprehending data and coming up with understanding of the data. 
In other words not just extracting analytics from the data but also synthesizing new knowledge from the data. 
Hence, we propose an architecture based on intuition to process big-data. We also propose a set of test-cases  to check the final performance of such a system, inspired by psychological studies of intuition. 

Rest of the paper is detailed as follows. Section 2 gives a brief description of some of the current knowledge on intuition. 
Section 3 draws from Section 2 and develops an architecture schema for the proposed system. 
Section 4 explains the proposed test-cases and Section 5 concludes the paper. 

\section{A Brief Note on Intuition}
	Intuition has been argued and accepted widely as a powerful ``human''�� trait \cite{bastick_82}. 
	As per the Oxford English Dictionary, intuition is ``the ability to understand or know something immediately, without conscious reasoning.". 
	The review paper by Hodgkinson etal \cite{hod_08} gives a thorough analysis of the state of the art understanding of intuition. 
	 Miller and Ireland discuss that intuition can be decomposed of two types of abilities, viz. automated expertise (an ability to jump to decision about action to be taken based on expertise gained through domain expertise) and holistic hunch (a less well-understood quick judgment through subconscious synthesis of information). 
	 Hence, one of the ways to implement wil be as follows. 
	 First of all a large number of domain specific patterns  are stored in the long term memory. And these patterns should not be stored like photographic memory, rather after being broken into cues and symbols \cite{chase_73}. And lastly these cues are linked through a relational graph which should not be fixed.
	
We can note the following salient features of intuition as has been observed and modeled by psychologists and neurologists. We note them so as to make sure these are modeled in the architecture that we shall present in Section~\ref{arch}. 
Hence, with each feature we shall also note what might be needed in order to implement that feature using machine. 
	\begin{itemize}
	\item Intuition is not very different from impulsive nature. Hence, to get a system to have intuitive abilities, the system should also be allowed to err. 
	It can be mentioned here that all AI experts (including Turing \cite{turing_50}) have suggested this very clearly that an AI system has to be allowed to err. 
	Hence, any system aiming to emulate intuition should also have a scheme to check the proposed action. 
	\item Intuition is gained from domain specific ``experience''. 
	We must note that ``experience'' does not mean just labeled data (as is the case with current phase of machine learning). ``Experience'' can be gathered from a range of sources which means the data will be in confusing format and sometimes (like in medical diagnosis retrieved from various hospitals) the labels might be contradictory.  Hence first of all we would need a sanity check while saving the data and the data should be saved in special format which aids cross-referencing. 
	Secondly, the learning phase has to change completely from the  learning schema of existing machine learning architecture. 
	\item It has been argued that humans keep on performing mental simulation to get the intuitive solution. 
	This is why the experts are good at it. 
	These mental simulations can be thought of as subconscious thought experiments. 
	Hence, the intuition machine need to have a daughter platform to keep running mental simulations in order to bolster itself. 
	\end{itemize}

\section{ICABiDAS Architecture \label{arch}}
Based on the discussion we presented in the previous section we develop an architecture in this section which can be used to model ``intuition'' into machine and thereby aid in not only extracting analytics from big-data but also in synthesizing newer interpretation of the data. 

\begin{figure}[htbp]
\includegraphics [scale=.3]{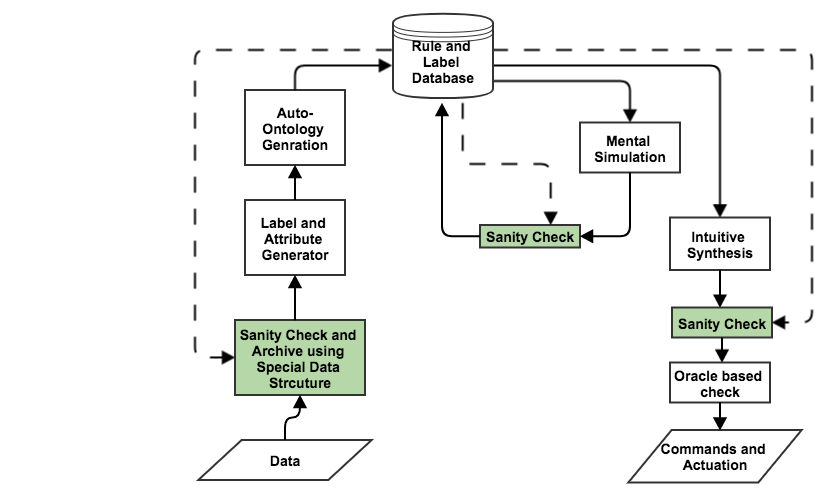}
\label{intu}
\caption{Here we show the proposed architecture. The blocks in light green depict the blocks where machine based filtering of data and interpretation is done. We call it ``sanity check''. }
\end{figure}

The architecture is presented in Figure~\ref{intu}. There are three major stages of the architecture. 
\\
The first phase is the input stage in which data from sensors and/or data-bases are sent into the system. 
As we have discussed the data might be in different formats and may have inherent glitches. 
Hence in this phase the data is passes through a ``sanity check'' block to make sure that the data has no anomaly. And then the data is converted into a universal way of representation. 
We propose to use a relational graph based representation to store data as relational information is one of the major reasons why human brain can process the amount of data it can. 
After extracting ontological information about the data the data and rules are stored in the central data-base\footnote{We are aware of the fact that human brain is a highly distributed system. However for the sake of simplicity we are using a centralized data and rule storage in our architecture.}. 
\\
The second stage is the mental simulation stage. 
This is a stage that can run in the background either on a regular basis (to synthesis knowledge about the existing dataset) or on demand (to generate intuitive action plan for a suggested problem). 
The extracted labels and rules in this phase are also passed through a sanity check block to make sure that nothing drastic and extremely counterintuitive gets stored into the central storage. 
\\
The third stage is the action stage which is used to synthesize new knowledge or action plan and given as output. 
The output from this passes through a sanity check again. 
This also need to pass through an oracle who can be a human expert. This is because the system will commit errors and if the system is a high-cost environment expert system it is better to pass the final action plan through a human oracle. 

As we discussed in the previous section, a major shift of paradigm we need to make so as to make an intuition engine work is to let it commit error. 
That is why we have introduced ``sanity check'' blocks in each stage of the schema.

\section{Suggested Test-cases}
There are a number of schema and architectures available in the open literature proposing different ways to handle big-data. 
However, without a firm set of test-cases it is of little use to the engineering and researching communities. 

We believe there are three phases to any new {\it architecture development}. 
In the first phase the architecture and the schema are designed. 
In the second phase the problem is understood and hence emerges a {\it set of test cases}. 
The schema and the desired performance is then followed by the third phase which is {\it implementation phase}.

In this section we shall present three test cases to test the intuitive ability of an implementation.  
As the field is undergoing aggressive research we shall endeavor to keep the test-cases as abstract as possible. 
And we shall also keep the test cases inspired by what has been used by psychologists. Hence we shall name the test-cases based on the psychological tests which has inspired them. 
As a case-study  we shall keep medical expert system in our mind and give examples from that domain. 

\begin{enumerate}
\item {\bf Westcott test based test-case:} One of the oldest tests of intuition is the Westcott’s test \cite{Westcott_68} in which a subject is presented with a problem with missing cues (so that a logical solution to the problem is impossible). 

This is a test to check the performance when sufficient cues are lacking. 
For example a medical expert system (MES) can be checked for its Intuitive ability (IA) by checking how well it can diagnose a disease when not all the diagnostic test results are not fed into the system. 

\item{\bf ACT test-case:} Another accepted test is the accumulated clues task (ACT) \cite{bowers_90}. In this the subject is presented with a list of words and one which can be linked to all other words. Its impossible to get to the right answer under time constraints, and hence it is believed to test intuition.

This is a test of predicting relation knowledge without enough data. For example an MES can be tested for its IA by checking if it can predict about the geographical location of the patients just from the diagnostic data (given it has enough data about patents from the surrounding region). 

\item{\bf  Waterloo Gestalt closure task based test-case:} Lastly there is the Waterloo Gestalt closure task \cite{ bowers_90} in which very ill defined images are presented to the subjects who are asked to identify them with a given time constraint. This tests the accumulated expertise kind of intuition which comes from domain expertise.

This is a test of determining the boundary conditions of a problem. 
For example an MES can be tested for its IA by checking if it can predict the estimated number of days a given patient may be hospitalised just from its diagnostic data. 
\end{enumerate}

The difficulty in defining what the intuitive level makes evaluating it all the more difficult. 
  On a side note it can be marked that time constraint is a common criteria while measuring intuition \cite{clark_05, hod_08}.

\section{Conclusion}
 In this paper we presented a schema to implement a bio-inspired architecture to manage big-data. 
 The major novelty is the inclusion of intuition which we believe is central to the way humans are able to process huge amount of sensory data and use them pertinently. 
 In addition to the schema we also present a set of test-cases which can be used to validate if a designed system has intuitive abilities and to what extent. 
 
 In our future work we shall implement social media comment analysis using our system and validate its working. \bibliographystyle{IEEEtran}
\bibliography{../ref}

%

\end{document}